# Leveraging Generative AI for Clinical Evidence Synthesis Needs to Ensure Trustworthiness

Gongbo Zhang[1], Qiao Jin[2], Denis Jered McInerney[3], Yong Chen[4], Fei Wang[5,6], Curtis L. Cole[5,7], Qian Yang[8], Yanshan Wang[9], Bradley A Malin[10], Mor Peleg[11], Byron C. Wallace[3], Zhiyong Lu[2], Chunhua Weng[1,\*], Yifan Peng[5,\*]

1. Columbia University, Department of Biomedical Informatics, New York, 10032, US
2. National Institutes of Health, National Library of Medicine, National Center for Biotechnology Information, Bethesda, 20894, US
3. Northeastern University, the Khoury College of Computer Sciences, Boston, 02115, US
4. University of Pennsylvania, Department of Biostatistics, Epidemiology and Informatics, Philadelphia, 19104, US
5. Weill Cornell Medicine, Department of Population Health Sciences, New York, 10065, US
6. Weill Cornell Medicine, Department of Medicine, New York 10065, US
7. Weill Cornell Medicine, Institute of AI for Digital Health, New York, 10065, US
8. Cornell University, Computing and Information Science, Ithaca, 14853, US
9. University of Pittsburgh, Department of Health Information Management, Pittsburgh, 15260, US
10. Vanderbilt University Medical Center, Department of Biomedical Informatics, Nashville, 37203, US
11. University of Haifa, Department of Information Systems, Haifa, 3498838, Israel
\* Corresponding author(s): Yifan Peng (yip4002@med.cornell.edu), Chunhua Weng (cw2384@cumc.columbia.edu)

**Abstract**

Evidence-based medicine promises to improve the quality of healthcare by empowering clinical decisions and practices with the best available evidence. The rapid growth of clinical evidence, which can be obtained from various sources, poses a challenge in collecting, appraising, and synthesizing the evidential information. Recent advancements in generative AI, exemplified by large language models, hold promise in facilitating the arduous task. However, developing accountable, fair, and inclusive models remains a complicated undertaking. In this perspective, we discuss the trustworthiness of generative AI in the context of automated evidence synthesis.

**Main**

Evidence-based Medicine (EBM) aims to improve healthcare decisions by integrating the best available evidence, clinical proficiency, and patient values. One widely utilized method for generating optimally accessible evidence is the systematic review of Randomized Controlled Trials (RCT). This involves identifying, appraising, and synthesizing evidence from relevant studies addressing a clinical question. While RCTs have been established as a reliable method for obtaining high-quality evidence for the safety and efficacy of medical products, their practical applicability is often curbed by cost constraints, ethical dilemmas, and logistical challenges. Consequently, some people have turned to real-world evidence, which is derived from real-world patients and observational data about them, such as electronic health records, insurance billing databases, and disease registries [1], for estimating treatment effects [2,3].

Clinical evidence synthesis can be defined as the act of grained analysis of evidence across publications to collect the most relevant evidence to a target study [4]. However, due to the rapid growth of the biomedical literature and clinical data, clinical evidence synthesis is struggling to keep up. In the meantime, the proliferation of misinformation or biased/incomprehensive summaries resulting from unreliable or contradicting evidence erodes the public's trust in biomedical research [5,6]. Given such concerns, we need to develop new and trustworthy methods to reform systematic reviews.

In recent years, dramatic advances in Generative AI, particularly Large Language Models (LLMs), have demonstrated a remarkable potential for assisting in systematic reviews [7]. Recently, LLMs have been explored to summarize meta-analyses [8] and clinical trials [9,10]. Before the era of LLMs, AI methods were also deployed to extract evidential information [11–16] and retrieve publications on a given topic [17–19]. However, achieving trustworthy AI for clinical evidence synthesis remains a complex undertaking. In this perspective, we discuss the trustworthiness of generative AI and the associated challenges and recommendations in the context of fully and semi-automated clinical evidence synthesis (Figure 1).

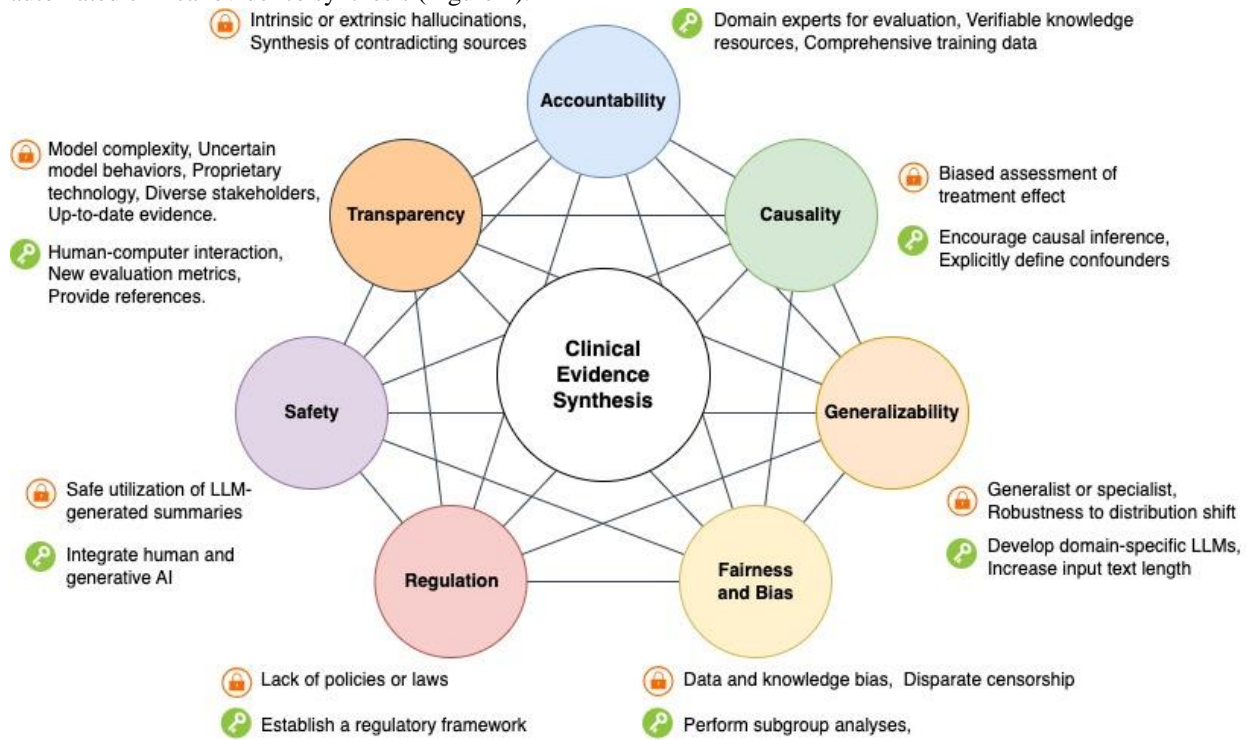

Figure. 1. Challenges and recommendations to achieve trustworthy evidence synthesis.

**Accountability**

In the context of summarizing clinical evidence, the accountability of an LLM refers to the model's ability to faithfully summarize high-quality clinical trials and the corresponding meta-analyses. In the context of health and human services, faithful or factual AI refers to systems that generate content that is factually accurate so that it is exchangeable. [20]. When using the term trustworthiness, we emphasize not only the factualness or faithfulness but also the system's reproducibility. While LLMs can generate semantically meaningful and grammatically correct textual outputs, these models can potentially yield factually incorrect outcomes [8]. These discrepancies can be classified as intrinsic or extrinsic hallucinations [21]. Intrinsic hallucinations refer to cases where "the generated output contradicts the input source." By contrast, extrinsic hallucinations occur when the generated output can "neither be supported nor be contradicted by the source."

The problem of factually incorrect outputs is partially due to differences in the sources of evidence and the resulting summary [21]. Particularly in evidence synthesis, the findings of the included clinical trials may not inherently align with the outcomes of the meta-analysis. Systematic reviews are designed to offer a statistical synthesis of the results of eligible clinical studies, rather than merely replicating them verbatim [22]. It is also plausible that crucial information from an included study might be omitted from the synthesis if perceived as offering low-quality evidence. Given the problems that end-to-end synthesis/summarization models have with aggregating information from contradictory sources [23], automatically generated summaries based solely on included clinical trials—without considering the meta-analysis—may not be reliable. Since the process of systematic review consists of multiple steps, directly utilizing LLMs as an end-to-end pipeline may increase the risk of incorrect outcomes due to errors in upstream tasks, such as searching, screening, and appraisal. The LLMs as an end-to-end system are also more challenging to understand than a system dedicated to a single subtask of the process.

Another factor contributing to the disparity between input evidence and summary is the accessibility of the included clinical studies, which is constrained by the copyrights of specific journals where the studies are published. For instance, if LLMs are instructed to summarize ten studies, yet only eight are publicly accessible and included as input, the models are compelled to generate speculative content to compensate for the missing sources of evidence. Therefore, comprehensive training data is critical for developing accountable models to summarize clinical evidence. All the information in the targeted summary should be referenced or derived from the included clinical trials, which are considered high-quality evidence. Additionally, reliable implementation of automatic meta-analysis workflow is critical to assure the correctness of statistically synthesized effect measures and their corresponding accuracy.

Further, it is important to determine how LLMs should be evaluated in evidence synthesis. While multiple automatic evaluation metrics have been developed for assessing AI-generated summaries, they do not correlate strongly with human expert evaluations concerning factual consistency, relevance, coherence, and fluency [8–10,24]. Thus, there is a need for a complete automatic evaluation protocol for evidence synthesis, as well as advanced research in developing evaluation metrics that better correlate with human judgments [25,26]. These automatic evaluation metrics should serve as a tool for complementing manual evaluations from domain experts and not replace them, especially considering the high-stakes nature of the task.

Beyond the above issues, parametric knowledge bias is a conspicuous problem [27]. This occurs when models depend on their intrinsic parametric knowledge, which is built up during training, rather than the information provided in the input source when generating summaries. To alleviate this issue, retrieval-augmented generation [28], which has been proposed to retrieve references and incorporate portions of these references into the completion, can be employed to enhance the accuracy and coherence of prompt completion. It enables healthcare providers to verify the evidence synthesis against referenced studies and assess the quality of the references [29].

**Causality**

Estimating treatment effects is important for informing clinical decisions. For decades, RCTs have been considered a gold standard, where patients are usually randomly assigned to the treatment or control group so that observable confounding factors are evenly distributed in the two groups. However, there are several limitations of RCTs that have yet to be recognized. For example, it is unethical to explore all possible outcomes if there is a risk of harm to a patient involved in the treatments. Also, statistically meaningful RCTs are challenging to administer for rare diseases. Thus, as a supplement to the evidence, observational data has been explored for estimating treatment effects [2]. Yet, despite the promising abundance of records in observational studies, the treated and untreated groups may not be directly comparable because of confounding factors. Even more problematic is that certain confounding factors, such as a patient's familial situation or economic status, may not be observable. As such, causal inference remains a challenge in leveraging observational data.

LLMs have demonstrated numerous breakthroughs in making inferences based on patterns (e.g., writing poems or source code [30]) and hold promise in assisting causal inference tasks by identifying confounding factors or generating descriptions for causal relationships. However, it is notable that LLMs may still "exhibit unpredictable failure modes [in causal reference]" [31]. Furthermore, there is ongoing discussion regarding whether LLMs truly perform causal reasoning or mimic memorized responses. As such, research needs to be conducted into how to best characterize LLMs' capacity for causality and understand their underlying mechanisms.

**Transparency**

In healthcare, it is critical that the systems are transparent due to their proximity to human lives and that patients understand how clinicians use these recommendations. In the context of LLMs, the challenges are unique, compared with traditional AI approaches, because of the complexity of the models, capability unpredictability [32], and proprietary technology.

Modern LLMs are typically based on neural models, which consist of multiple layers of interconnected neurons, and the relationship between input and output can be highly complex. It is thus challenging to understand how a specific summary is generated based on the input information, which poses a challenge in building transparent systems in evidence synthesis. This complexity is exemplified by the *exposure bias* problem, which refers to the difference in decoding behavior between the training and inference phases. It is common practice to train models to generate the next token conditioned on ground-truth prefix sequences. However, during inference, the model generates the next token based on the historical sequences it has generated. The model can be potentially biased to perform well only when conditioned on truth prefixes. This is also referred to as *teacher forcing*. This discrepancy can trigger progressively more erroneous generation, particularly as the target sequence lengthens. While the problem was identified almost a decade ago [33], it is still unclear to what extent exposure bias affects the quality of model output [34].

In recent years, researchers in AI and human-computer interaction (HCI) have focused on developing and evaluating different approaches to achieve transparency in clinical AI systems, such as model training techniques, standard protocols for documenting the data and model training process [35–37], techniques that explain the confidence level of AI predictions in ways consistent with how clinicians weigh uncertainty in clinical decision making and explain clinicians' decisions to patients [38]. Additional work focused on establishing community guidelines that are informed by human-centered design principles for such applications that potentially influence health behaviors [39]. These prior studies established a solid foundation for addressing the challenges generative AI brings and, therefore, can be valuable for guiding the development of AI-generated summaries.

Regarding EBM, we need to create teams of diverse stakeholders, including, at a minimum, patients, healthcare practitioners, and policymakers. Representing and supporting their needs is critical to ensuring that the generative AI research community develops meaningful and respectful technologies. For instance, clinical studies summarized for patients should prioritize readability and comprehensibility. In contrast, summaries intended for healthcare practitioners should provide sufficient detail to support trustworthy decision-making. Additionally, the versions crafted for policymakers should highlight potential risks to the synthesis process and discuss their broader implications.

Finally, developing models with baked-in structures is crucial to achieving transparency. For example, Saha *et al.* [40] use a systematically organized list of binary trees that symbolically represent the sequential generative process of a summary from the source document and highlight significant pieces of evidence that influence the synthesis, and Ramprasad *et al.* [10] separate conditions, interventions, and outcomes in input RCTs to yield an aspect-wise interpretable summary. This also means that transparency is not an afterthought but a significant element that's diligently crafted and assessed with the expertise of domain professionals.

**Fairness**

LLMs offer significant benefits for addressing biases in clinical trials and enhancing research inclusivity. For example, LLMs can process and analyze vast amounts of data from diverse populations. This ensures that the findings are more representative and applicable to a wider range of patient groups. Moreover, LLMs can process information in multiple languages, facilitating accessibility to clinical trial data and evidence synthesis for researchers and practitioners worldwide. However, it is important to note that while some LLMs can generate documentation on par with clinicians, there is evidence that LLMs can also propagate biased results [41]. For example, there may be a need for LLMs to better capture the demographic diversity of clinical conditions to prevent generating clinical vignettes that perpetuate stereotypes about demographic presentations [42]. The presence of biases in the LLM's training data can also perpetuate or amplify these biases in its analysis and summaries. In the healthcare domain, this is particularly concerning, as biased data can lead to unfair and harmful outcomes. Moreover, LLMs may not fully understand the cultural, social, and ethical contexts of clinical trials, potentially leading to oversimplified or inappropriate conclusions that might not be fair or applicable to all patient groups. Therefore, the development, deployment, and use of generative AI should strive for fairness.

While quality assurance for clinical studies is orthogonal to model training and inference, the underlying bias can be amplified in the synthesized evidence summaries based on the clinical studies. Subsequently, the spread of mainstream misinformation and the fast publication of invalid evidence has dramatically undermined the public's trust in biomedical research [43]. Even though the evidence synthesis task might be less prone to biases in evidence-based medicine, it is still crucial to remain cautious. One possible way to mitigate these issues is to construct prompts (i.e., instructions to LLMs) explicitly requesting to avoid bias. However, such a bias-free prompt is unlikely to be common practice because biases can significantly influence the overall quality of the model development and may be present more broadly and deeply within the LLMs.

In the context of real-world evidence, it has been found that machine learning (ML) models may perform poorly on under-represented groups [44,45]. Such situations where a patient group is under-tested compared to others are called *disparate censorship*. For training supervised learning models, patient outcomes are labeled based on diagnostic test results. While medical records list diseases that patients have been diagnosed with, the absence of a diagnosis does not mean the patients do not have such a disease. Assuming undiagnosed patients are healthy can lead to the development of biased ML models that produce incorrect summaries, particularly for patient groups that have limited access to healthcare [44,45]. Clinical decisions based on biased generations, *e.g.,* evidence summaries, can further harm the already under-tested population [46]. The research community needs to focus on assessing the LLMs to ensure they do not exhibit biases, discrimination, and stigmatization towards individuals or groups [42,47].

**Model Generalizability**

Another crucial component of achieving trustworthiness for generative AI is generalizability. The models must behave reliably and reproducibly while minimizing unintentional and unexpected consequences, even when the operating environment differs from the training environment.

While most popular LLMs are trained using diverse text resources and have demonstrated considerable proficiency, they need a deeper understanding of specialized fields, particularly in the clinical domain. For instance, domain-specific LLMs may technically comprehend clinical knowledge, such as the UMLS terminology, more than their generalist counterparts [48,49]. To this end, we should prioritize constructing domain-specific LLMs and benchmarks [50], as generic benchmarks are no longer of primary importance in evidence-based medicine. So far, multiple LLMs have been developed for medical applications and open-sourced, such as BioBERT, ClinicalBERT, PubMedBERT, BioMedLM, and GeneGPT [51–55]. In addition, many companies adapted their close-sourced models for medical applications, such as GPT-3.5, GPT-4 by OpenAI, and MedPaLM by Google [56–58]. These LLMs were pre-trained on datasets that do not include the most recent knowledge and information in the medical domain, which is a fast-evolving field. As such, developing domain-specific models is an on-going process and a high priority.

Another challenge for generalizability in evidence synthesis is the ability to process long inputs. Even though there is a trend toward increasing the maximum length of input tokens, the extended context window may still need to be adequately long to encompass all clinical trials or notes involved in an evidence summary. To combat this issue, several strategies have been proposed, such as chunking the document, hierarchical summarization, and iterative refinement by looping over multiple sections or documents [59,60]. A potential solution for summarizing multiple clinical trials that cannot fit in a single prompt is maintaining historical interaction records. These records can be used to process clinical trials in batches. Additionally, exploring hierarchical summarization mechanisms could enable synthesizing evidence from concise summaries of the clinical trials, rather than directly from the trials themselves.

**Data Privacy and Governance**

There are numerous concerns over how an LLM may share and consume patient information. However, it is still being determined if patients would consent to use information about them in an LLM - particularly if they are not informed about, or unable to comprehend, what the LLM would be used for. In this respect, if patient information is to be relied upon, there need to be clear descriptions of anticipated uses - even if such a communication indicates that it is unknown what such services may be.

Even if patients consent to transfer information about them to an LLM, they may not wish to have potentially identifying or stigmatizing information about them disclosed to or retained by the LLM. While LLMs are designed to create probabilistic representations of the relationships between features, if not designed appropriately, they may memorize training instances supplied to them [61]. As such, when the LLM is queried, it may be possible for the end user to determine the training data, which could have legal implications [62]. Even if individual-level records cannot be recovered from an LLM, it may also be possible that a patient can be detected as a contributor to the training data [63]. Such membership inferences could be problematic if, for instance, the LLM has been fine-tuned on patients diagnosed with a sensitive clinical phenomenon, such as a sexually transmitted disease. There have been various investigations into incorporating formal privacy principles, such as differential privacy[62], into the construction and use of LLMs; however, it is currently unclear how such noise-based mechanisms influence clinical trial synthesis capabilities. In this respect, it is critical to continue research into how best to train and apply LLMs in a manner that respects the privacy of the patients upon whom the technology is based.

**Patient Safety**

Safety also remains a pressing concern when utilizing clinical evidence summarized by LLMs, as any inaccuracies could have far-reaching implications for high-stakes healthcare decisions. To mitigate these risks, generative AI should come equipped with protective measures that facilitate a backup plan in case of any complications. As we continue to advance in AI, the goal is for LLMs to transform from simple tools to collaborative partners operating in sync with human experts. While using LLMs to quickly analyze and synthesize large amounts of evidence could expedite the evidence synthesis process, a secure and reliable architecture is essential to ensure that expert meta-analyzers can trust these AI-generated summaries.

One possible solution revolves around a synergy between humans and generative AI, with a focus on their mutual safety. This collaborative framework would involve human experts frequently reviewing and offering feedback on summaries produced by LLMs, thereby resulting in iterative improvements. In this way, humans and generative AI can develop trust in each other and continue to evolve concurrently, enhancing their teamwork. We believe this integration will enhance capabilities beyond what humans alone can achieve, introduce new capacities, and, crucially, embody "high confidence." Accordingly, they should exhibit reliable or predictable performance, demonstrate tolerance towards environmental and configuration faults, and maintain resilience against adversarial attacks.

**Lawfulness and Regulation**

Finally, generative AI should unequivocally adhere to all pertinent laws and regulations, including international humanitarian and human rights laws, a cardinal principle that carries particular significance when applying generative AI for evidence synthesis. This entails recognizing the nuanced legal landscape governing AI deployment, which may diverge from one state or country to another. Thus, it becomes imperative to construct a comprehensive legal framework addressing the accountability associated with actions taken and recommendations made by AI systems.

Under the General Data Protection Regulation (Article 9), collecting and processing sensitive personal data is subject to strict regulations [64]. The North Atlantic Treaty Organization (NATO) has established guidelines promoting the responsible utilization of AI, including a principle emphasizing that AI applications will be created and employed in compliance with both domestic and international legal frameworks [65]. Recently, the United States Congress has conducted a series of testimonies and hearings involving AI companies and AI researchers to address matters pertaining to AI regulation and governance [66,67]. The European Union has also initiated the process of creating regulations regarding developing and using generative AI [68]. In alignment with these movements, LLMs for evidence synthesis and beyond must also be developed with these legal challenges in mind to safeguard patients, clinicians, and AI developers from any unintended repercussions.

**Conclusion**

Generative AI for clinical evidence synthesis has already made a positive impact and is set to continue doing so in a way we cannot imagine. However, it is equally vital that risks and other adverse effects are properly mitigated. To this end, building generative AI systems that are genuinely trustworthy is crucial. This document outlines several directions that efforts could take to achieve trustworthy generative AI, ideally working in unity and overlapping in their functionality.

While this document primarily targets evidence synthesis, some principles could prove beneficial in other domains. The present moment necessitates the creation of a culture of Trustworthy AI within the JBI community, enabling the benefits of AI to be fully realized within our healthcare system.

**Acknowledgments**

This research was sponsored by the National Library of Medicine grant R01LM009886, R01LM014344, R01LM014306, and the National Center for Advancing Clinical and Translational Science award UL1TR001873. It was also supported by the NIH Intramural Research Program, National Library of Medicine.

**Contributorship**

Study concepts/study design, all authors; manuscript drafting or manuscript revision for important intellectual content, all authors; approval of the final version of the submitted manuscript, all authors; agrees to ensure any questions related to the work are appropriately resolved, all authors; literature research, all authors; and manuscript editing, all authors.

**Conflict of Interest**

None.